\documentclass{WileyMSP-template}

\usepackage{graphicx}
\usepackage{xcolor}
\usepackage{subcaption}
\usepackage{textcomp}
\usepackage{multimedia}
\usepackage{color,soul}

\renewcommand\hl[1]{#1} 

\begin{document}

\pagestyle{fancy}
\rhead{\includegraphics[width=2.5cm]{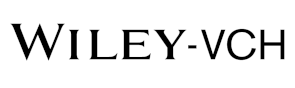}}

\title{Novel Design of 3D Printed Tumbling Microrobots for \textit{in vivo} Targeted Drug Delivery}

\maketitle

\author{Aaron C. Davis}
\author{Siting Zhang}
\author{Adalyn Meeks}
\author{Diya Sakhrani}
\author{Luis Carlos Sanjuan Acosta}
\author{D. Ethan Kelley}
\author{Emma Caldwell}
\author{Luis Solorio}
\author{Craig J. Goergen}
\author{David J. Cappelleri*}


\dedication{}

\begin{affiliations}
Aaron C. Davis, Emma Caldwell, Prof. Craig J. Goergen, and Prof. David J. Cappelleri\\
School of Mechanical Engineering 
Purdue University 
585 Purdue Mall, West Lafayette, IN 47907, USA\\
Email Address:~\{davi1381,caldwe64,cgoergen,dcappell\}@purdue.edu 

Siting Zhang, Adalyn Meeks, Diya Sakhrani, Luis Carlos Sanjuan Acosta, D. Ethan Kelley, Prof. Luis Solorio, Prof. Craig J. Goergen, and Prof. David J. Cappelleri\\
Weldon School of Biomedical Engineering
Purdue University 
206 S Martin Jischke Dr., West Lafayette, IN 47907, USA\\
Email Address:~\{zhan3465,meeks18,ddsakhra,lsanjuan,kelle278,lsolorio,cgoergen,dcappell\}@purdue.edu \\
Prof. Luis Solorio, Prof. Craig J. Goergen, and Prof. David J. Cappelleri\\
Purdue Institute for Cancer Research\\
201 South University Street, West Lafayette, IN 47907, USA\\
\textbf{Funding:} This work was supported by NIH Award 1U01TR004239-1 and the Purdue Institute
for Cancer Research Pilot Grant 2023-24 Cycle 2 Award.  Funding sources did not have any role in the study design; in the collection, analysis and interpretation of data; in the writing of the report; and in the decision to submit the article for publication.

\textbf{Ethics approval:} All animals and methods used are in compliance with Purdue University's Institutional Animal Care and Use Committee (IACUC) under protocol number \#2002002016.

\textbf{Competing interests:} Craig J. Goergen is a paid consultant of FUJIFILM VisualSonics Inc. VisualSonics had no role in the collection or analysis of these data.
\end{affiliations}



\begin{abstract}
This paper presents innovative designs for 3D-printed tumbling microrobots, specifically engineered for targeted \textit{in vivo} drug delivery applications. The microrobot designs, created using stereolithography 3D printing technologies, incorporate permanent micro-magnets to enable actuation via a rotating magnetic field actuator system. The experimental framework encompasses a series of locomotion characterization tests to evaluate microrobot performance under various conditions. Testing variables include variations in microrobot geometries, actuation frequencies, and environmental conditions, such as dry and wet environments, and temperature changes. The paper outlines designs for three drug loading methods, along with comprehensive assessments thermal drug release using a focused ultrasound system, as well as biocompatibility tests. Animal model testing involves tissue phantoms and \textit{in vivo} rat models, ensuring a thorough evaluation of the microrobots' performance and compatibility. The results highlight the robustness and adaptability of the proposed microrobot designs, showcasing the potential for efficient and targeted \textit{in vivo} drug delivery. This novel approach addresses current limitations in existing tumbling microrobot designs and paves the way for advancements in targeted drug delivery within the large intestine.
\end{abstract}

\section{Introduction}
\hl{The harmful side-effects and small therapeutic windows of drugs have been an ongoing problem in modern healthcare particularly in areas such as cancer and autoimmune diseases. In recent years, targeted drug delivery has shown the potential to revolutionize disease treatment by delivering therapeutic agents directly to the affected site while minimizing off-target effects. This approach enhances drug efficacy, reduces side effects, and improves patient outcomes. 
}

\hl{Mainstream approaches for targeted delivery have mostly focused on the use of nanoparticles and nanocarriers, such as liposomes, nanobubbles, DNA nanostructures, and micelles~\mbox{\cite{Tewabe2021, Evans2016a, Kumar2024, Delaney2022a, Rahim2021, Ashique2021, Rosenblum2018}}. These approaches--which primarily target tumors--are introduced to the body via the bloodstream. Encapsulation of a drug prolongs its life and reduces absorption which allows for greater accumulation at the target site. These carriers can be designed to include responsive components, where the payload release from the particles or carriers can be triggered via external stimuli such as light, heat, or magnetic fields~\mbox{\cite{Rahim2021}}, or by natural processes associated with the diseased state such as pH~\mbox{\cite{Galvin2012}}. Chemical modifications or physical design of the particles can also facilitate increased accumulation at the diseased site~\mbox{\cite{Galvin2012}}. 
}

\hl{Some regions of the body are not conducive to nanoparticle drug delivery. The release of drugs on demand at specific locations in the large intestine is difficult due to the large intestine's nature to quickly absorb or expel most material. For treatments of ulcerative colitis and Crohn's disease, targeted drug delivery is crucial to optimizing therapeutic outcomes without increasing systemic doses~\mbox{\cite{Liu2022, Ceylan2019c, Gurney1996, Gurney2002}}. Microrobotic drug delivery is a promising solution in this area.
}

Following initial results from the late 2000s~\cite{Li2009,Yesin2006,Abbott2009a,Kim2008}, the last decade has brought a surge of interest in leveraging microrobotic systems to increase patient health through the use of targeted drug delivery and minimally invasive interventions~\cite{Liu2022, Jang2019,  Park2019c, Ahmad2022, Son2020a, Yim2013a, Luo2018a, Soto2020a}. Microrobots present a promising avenue for improving therapeutic outcomes in targeted drug delivery due to their miniature size, enabling navigation through intricate biological pathways and delivering drugs with unprecedented precision~\cite{Nelson2010b, Shen2023}. 

Advancements in microrobotics have included the integration of multiple propulsion mechanisms, such as magnetic~\cite{Meeker1996, Ishiyama2001, Pawashe2009}, optical~\cite{Sul2006}, acoustic~\cite{Kagan2012, Ahmed2015}, and chemotactic~\cite{Mano2005, Solovev2009} methods. This enables controlled and directed movement within the human body~\cite{Jang2019, Chen2022, Ahmad2022}. Magnetic actuation, particularly, has gained interest for providing external control over microrobots, ensuring precise navigation and localization~\cite{Zhu2022, Wang2023}. Following trends in other fields, modern 3D printing technologies have proven to be useful even at the small scale of microrobotics~\cite{Wei2024, Huang2015a, Tan2023, Tan2024}. This technology enables exploration of 3D geometries and materials to enhance drug payload capabilities.

While the large intestine serves as a conducive entry point for microrobots in minimally invasive procedures it poses a complex environment for drug delivery, with intricate folds, varying diameters, and dynamic fluid motion. 
Tumbling microrobotic systems show particular promise in navigating the tortuous and confined spaces of the large intestine~\cite{Niedert2020}, enabling the precise delivery of therapeutic agents to localized regions. 
Despite promising advances, the field faces persistent challenges, notably the limited drug payload capacity due to the small size of microrobots. This constraint restricts the volume of therapeutic agents transported, preventing clinically significant doses. Overcoming this requires intricate design and advanced materials to enhance payload capacity, ensuring the feasibility of microrobotic drug delivery.  The challenge is creating microrobots with controlled drug release mechanisms, potentially triggered by external stimuli or physiological cues. Integrating smart, responsive materials~\cite{Tan2021b} is essential for precise drug release control, overcoming existing constraints. Real-time imaging within the large intestine is also necessary to monitor microrobot behavior. This requires high-resolution real-time feedback on microrobot movement, drug release, and tissue interactions, achievable through advanced imaging technologies such as endoscopes or ultrasound systems.


This paper addresses these challenges by introducing a magnetic tumbling microrobot ($\mu$TUM) that uses 3D printing technologies to substantially enhance payload capabilities, ensuring the administration of therapeutically consequential doses. The ($\mu$TUM) integrates thermally responsive materials into microrobot structures to actualize precise on demand drug release.\hl{ Drug release is triggered by localized heating using a focused ultrasound beam similar to the release that has been shown with some nanocarrier-based drug delivery systems~\mbox{\cite{Bachu2021, Evans2016a, Zhou2014a, Dromi2007a}}.} This work also underscores the integration of advanced imaging technologies, ensuring real-time monitoring of microrobot locomotion within confined spaces of the large intestine. Our research aims to expand the domain of microrobotic drug delivery, setting the stage for more effective and targeted therapeutic interventions in gastrointestinal healthcare.

\begin{figure}
    \centering
    \includegraphics[height=2.5in]{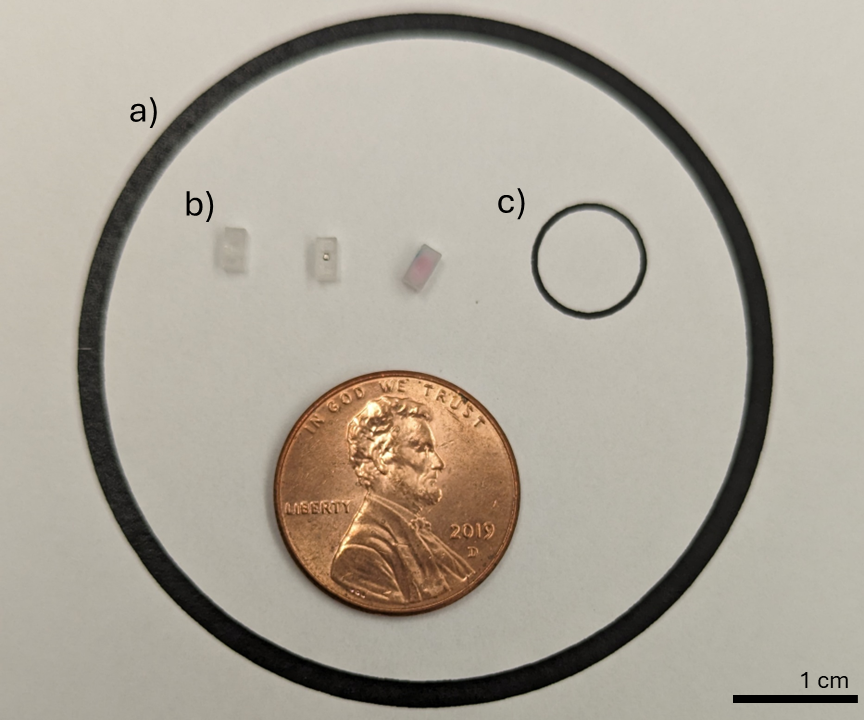}
    \caption{Magnetic tumbling microrobots ($\mu$TUMs) for \textit{in vivo} targeted drug delivery. a) The 50~mm average diameter of a human colon~\cite{Helander2014}. b) Top port $\mu$TUM designs from left to right as printed, with magnet inserted, and payload filled and capped. c) The 8.5~mm average diameter of a rat colon.}
    \label{fig:Coin}
\end{figure}


\section{Materials and Methods}
\subsection{Design Overview}
\begin{figure}[ht]
    \centering
    \begin{subfigure}[t]{0.32\linewidth}
        \centering
        \includegraphics[width=\linewidth]{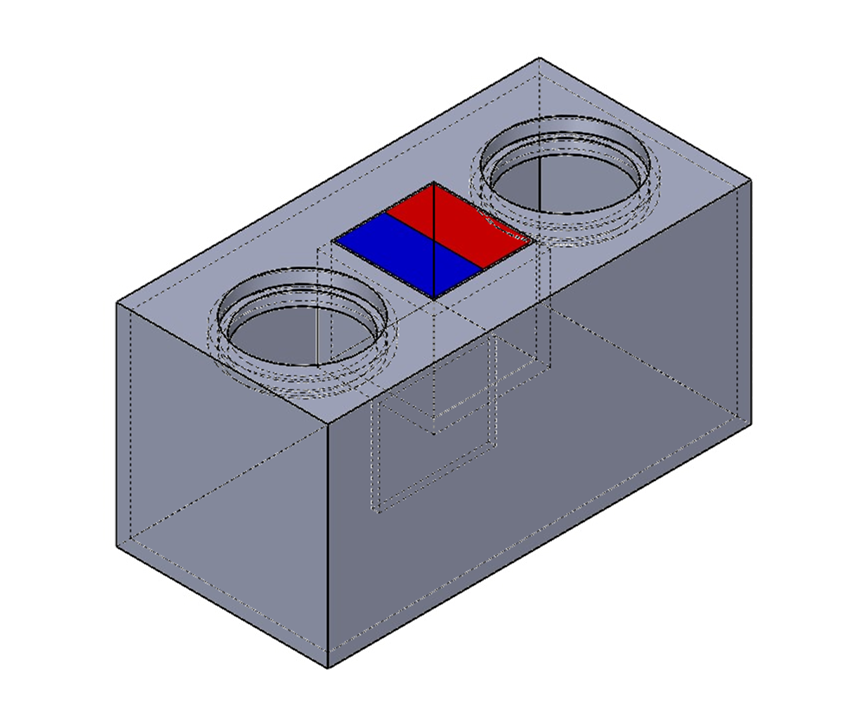}
        \caption{TP: Top Ports Design}
        \label{fig:TopPort}
    \end{subfigure}
    \hfill
    \begin{subfigure}[t]{0.32\linewidth}
        \centering
        \includegraphics[width=\linewidth]{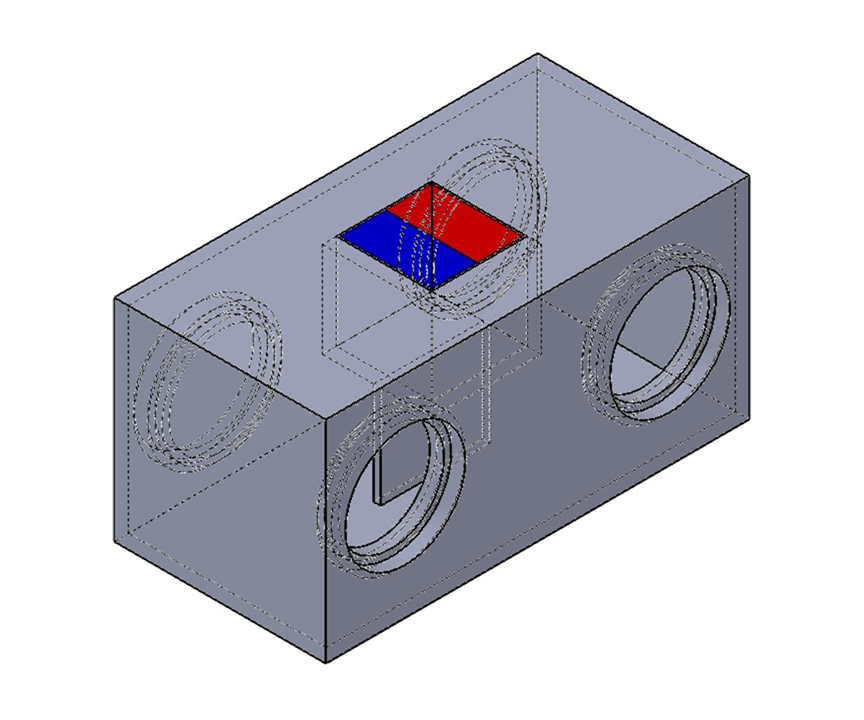}
        \caption{SP: Side Ports Design}
        \label{fig:SidePort}
    \end{subfigure}
    \hfill
    \begin{subfigure}[t]{0.32\linewidth}
        \centering
        \includegraphics[width=\linewidth]{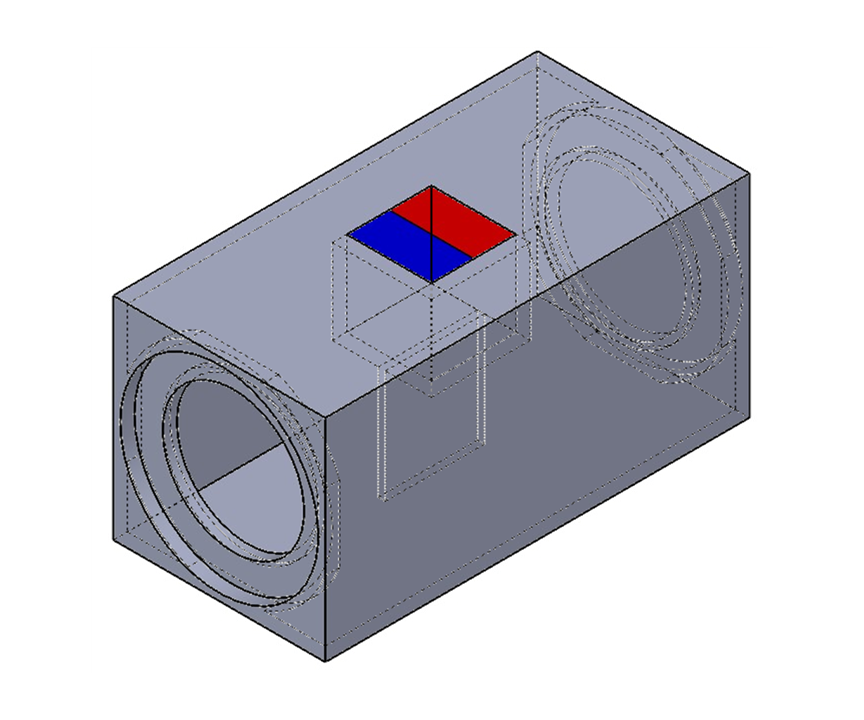}
        \caption{EP: End Ports Design}
        \label{fig:EndPort}
    \end{subfigure}
    \caption{3D printed magnetic $\mu$TUM designs with different port configurations and geometry for controlled drug release.
    }
    \label{fig:Designs}
\end{figure}

Under magnetic control microrobots are actuated by the magnetic force, $\vec{F}_m$, and the magnetic torque, $\vec{T}_m$. These are found using the following equations with the magnetic moment of the microrobot, $\vec{m_r}$, and the applied magnetic field, $\vec{B}$:
\[\vec{F}_m = (\vec{m_r}\cdot \nabla)\vec{B} \]
\[\vec{T}_m = \vec{m_r} \times \vec{B} \]
The microrobots presented here incorporate permanent micro-magnets into their structure, allowing them to respond to external magnetic fields. At the scale of these microrobots, the magnetic torques generated by a permanent magnet or electromagnetic coil are much larger--often several orders of magnitude--than the magnetic forces generated~\cite{sitti2017mobile}. This is one of the reasons tumbling is such an effective method of locomotion for microrobots. Tumbling motion is also effective in reducing the effect of the adhesive forces of the substrate on the microrobot such as those present inside the human body. When exposed to a permanent magnet, the magnetic force will pull the microrobot towards the magnet, and the magnetic torque will tend to align the microrobots with the field of the magnet. When the permanent magnet is rotated, the microrobot will follow the rotation, causing a tumbling motion. 

Due to their small size, mobile microrobots are specifically useful for navigating through delicate biological pathways with precision, facilitating access to remote or confined anatomical locations. Two-photon polymerization (TPP) 3D printing technology has been shown effective for the fabrication of microrobots ideal for minimally invasive interventions due to their enhanced maneuverability in narrow spaces \cite{Song2022,Adam2021c,Li2021,Rajabasadi2021a}. However, for applications like colon microrobotics, larger microrobots are more advantageous due to the colon's accessibility and size, requiring a shift to high precision stereolithography (SLA) 3D printing when the printing times required for TPP become impractical for larger scales.
Using SLA fabrication, we built 3~mm long microrobots for use in rat colons. Figure~\ref{fig:Coin} shows a sample $\mu$TUM design along with an U.S. penny and circles indicating the diameter of human and rat colons, respectively. \hl{For drug delivery in porcine or human colons, even larger robots capable of carrying multiple payloads or sensors could be explored}. The 3~mm long microrobots provide increased payload capacity, allowing for the carrying of a more substantial quantity of therapeutic agents. This is advantageous in applications where a higher drug dosage or a larger payload is required to achieve therapeutic efficacy. The larger physical dimensions of the 3 mm $\mu$TUMs also provide a more substantial and visually detectable entity. This increased size contributes to improved visibility under various imaging modalities, facilitating easier tracking and monitoring during operations or experiments. This is particularly important when using microrobots in larger animals as the resolution of ultrasound imaging is inversely proportional to the penetration depth~\cite{Szabo2004}. 

Loading microrobots with drugs involves incorporating drug payloads into the microrobot's structure in a manner that ensures controlled release at the desired location. To increase the drug capacity, we designed microrobots with a hollow cavity providing a space to encapsulate the drug payload. This cavity is then filled with a drug solution and then sealed using a temperature sensitive wax to prevent premature release and opened when thermally triggered \hl{using a focused ultrasound system}, allowing controlled drug release.

The design volume of the interior cavities of the $\mu$TUMs is 3~$\mu$L. The size and location of the opening to the drug cavity can be adjusted to affect the drug release profile in both passive and active drug release conditions and are explored here, as shown in Fig.~\ref{fig:Designs} and Table~\ref{tab:TUMparam}. This method balances control over the drug release profile and drug loading capacity.

\begin{table}
    \caption{$\mu$TUM design parameters}
    \centering
    \begin{tabular}{llll}  
    \hline
                                & \textbf{Top Ports (TP)}            & \textbf{Side Ports (SP)}           & \textbf{End Ports (EP)} \\\hline
            \textbf{$\mu$TUM Size}    & 3 x 1.4 x 1.4 mm$^3$ & 3 x 1.4 x 1.4 $mm^3$& 3 x 1.4 x 1.4 mm$^3$ \\
        \textbf{Port Count}     & 2                             & 4                             & 2 \\
        \textbf{Port Diameter}  & 750 \textmu m                 & 750 \textmu m                 & 1000 \textmu m \\
        \textbf{Internal Volume}& 5 \textmu L                   & 5 \textmu L                   & 5 \textmu L \\
        \textbf{Drug Cavity Volume}& 3 \textmu L                   & 3 \textmu L                   & 3 \textmu L \\
        \textbf{Magnet Size}    & 500 x 500 x 500 \textmu m$^3$& 500 x 500 x 500 \textmu m$^3$ & 500 x 500 x 500 \textmu m$^3$\\ 
\hline
    \end{tabular}
    \label{tab:TUMparam}
\end{table}

\subsection{Fabrication Procedures}

\subsubsection{Microrobot Body}
The microrobot housings are fabricated using a Form3+ (FormLabs) SLA 3D printer using Formlabs clear FLGPCL04 resin with 25~$\mu$m layer height and standard print settings. After printing, the microrobots are rinsed in isopropyl alcohol (IPA) for 10 minutes before cleaning in a sonication bath with fresh IPA for 20 minutes. 500~$\mu$m cube nickel-coated NdFeB permanent magnets (C000650-10, SuperMagnetMan) were then manually inserted into each robot and glued using UV curable adhesive (Loctite 3926) which was cured using a UV flashlight.

\subsubsection{Microrobot Drug Payload}
\begin{figure} 
    \centering
    \includegraphics[width=1\linewidth]{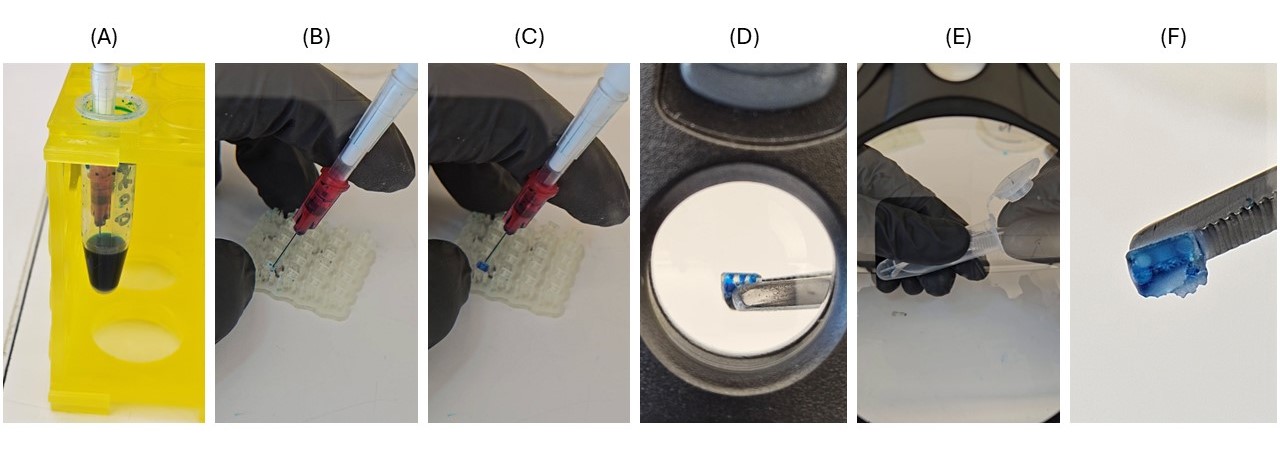}
    \vspace{-0.50 in}
    \caption{Steps for loading and coating $\mu$TUM-TP robots. These robots were loaded with a proxy drug solution consisting of blue food dye, 0.07\% Tween20, and MilliQ water.  They were sealed with a wax formulation of $w$ = 0.6 mass fraction of mineral oil. (A) Proxy drug is extracted using CELLINK 25 gauge high-precision blunt needle. (B) and (C): The solution is injected through the top ports of the robot. (D) The robot is held from its sides to prepare for the wax coating application.  (E) Coating the robot in the melted wax mixture. (F) The solidified wax coating on the robot after 2 minutes post coating.}
    \label{fig:LoadingDiagram}
\end{figure}

Figure~\ref{fig:LoadingDiagram} illustrates the process to load and wax coat a top port $\mu$TUM robot ($\mu$TUM-TP).  
First, 10 $\mu$L of a solution of interest is extracted with a pipette attached to a CELLINK 25 gauge high-precision blunt needle (Fig.~\ref{fig:LoadingDiagram}(A)). This solution can be a drug solution, a BSA solution, or a proxy drug solution (blue food dye, 0.07\% Tween20, and MilliQ). The CELLINK needle is placed through one of the ports and then the solution is released into the robot (Fig.~\ref{fig:LoadingDiagram}(B)). As the robot is being filled, the tip is carefully elevated (Fig.~\ref{fig:LoadingDiagram}(C)) to prevent loading bubbles inside the robot.

Once loaded, the robot is coated with a melted wax solution. 
The wax coating is a mixture of paraffin wax (Fisherbrand Histoplast LP Paraffin Wax) and mineral oil (Thermo Scientific Chemicals, Mineral Oil). Each component is weighed to achieve a desired formulation based on the mass fraction of mineral oil (see Sect.~\ref{sec:WaxFormulationMethod} for more information). Next, both components are combined and placed in a water bath at 80\degree C to melt the paraffin wax. Once all components are melted, it is removed from the water bath for storage or used immediately.
To coat the robot, the robot is held by its sides (Fig.~\ref{fig:LoadingDiagram}(D)) and quickly submerged \hl{one time} in the melted wax formulation resulting in a thin cap on top of the ports (Fig.~\ref{fig:LoadingDiagram}E). The wax cap will immediately start solidifying and after 2 minutes it is ready to be used for further experimentation (Fig.~\ref{fig:LoadingDiagram}(F)).  Excess wax can be scraped off during this 2 minute period, as needed.

\subsubsection{Phantom Model}

\begin{figure}
    \centering
    \includegraphics[height=2.5in]{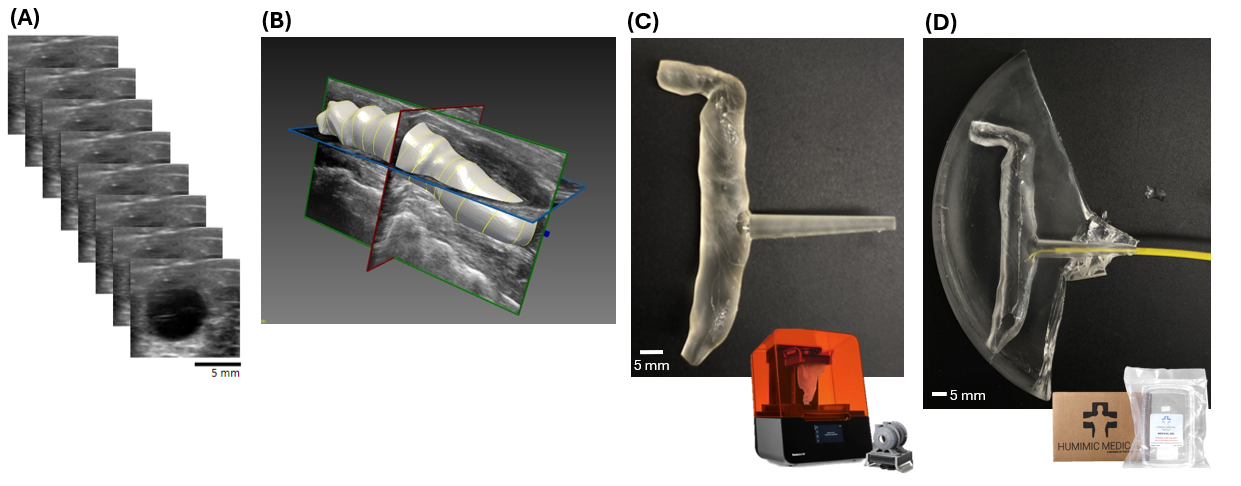}
    \caption{Rat colon phantom creation process.  (A) Short axis B-mode images of the inflated colon along the length of colon. (B) segmented 3D image of the colon using SimVascular with lofting. (C) 3D-printed positive of inflated colon using the Form3+ resin printer. (D) Inflated colon geometry in Humimic Medical’s Gelatin \#0 with inserted thermocouple.}
    \label{fig:createphantom2}
\end{figure}

To create the rat colon phantom model, the procedure to inflate a live rat colon (described in Sec.~\ref{sec:in_vivo_testing}) was followed and a volumetric (3D) scan of the inflated colon obtained.
To do so, short-axis B-mode ultrasound images (Fig.~\ref{fig:createphantom2}(A)) were acquired along the length of the colon using a step size of 0.10 mm. The images were then segmented using SimVascular (Fig.~\ref{fig:createphantom2}(B)) \cite{simvascular}. This segmentation of the Sprague-Dawley rat colon was used to create 3D printed positives using a 3D printer (Form 3+, Formlabs, Somerville, MA, USA). A 5 mm diameter cylinder, which served as an access point for a thermocouple device, was connected perpendicularly to the positive in the mid-distal region, as shown in Fig.~\ref{fig:createphantom2}(C). 
Gelatin \#0 from Humimic Medical was melted in a beaker on a hot plate until it reached an internal temperature of 48$^{\circ}$C. The melted Gelatin \#0 (Product Code: 852844007390, Batch Nos: HM02307171 and HM02406251) was then poured into a vessel containing the 3D-printed positive. The melted gel was transferred to a vacuum chamber, allowing bubbles to be removed from the gel. The gel was removed from the vacuum chamber after 20 minutes, before it began to set. The gelatin was allowed to cool completely before gently removing the 3D printed positive to form an intact negative of the colon. The added cylinder in the mid-distal region of the negative colon provides access to the thermocouple for precise temperature readings during testing (Fig.~\ref{fig:createphantom2}(D)).

\subsection{Locomotion Characterization Testing}
\begin{figure} 
\centering
\includegraphics[width=4in]{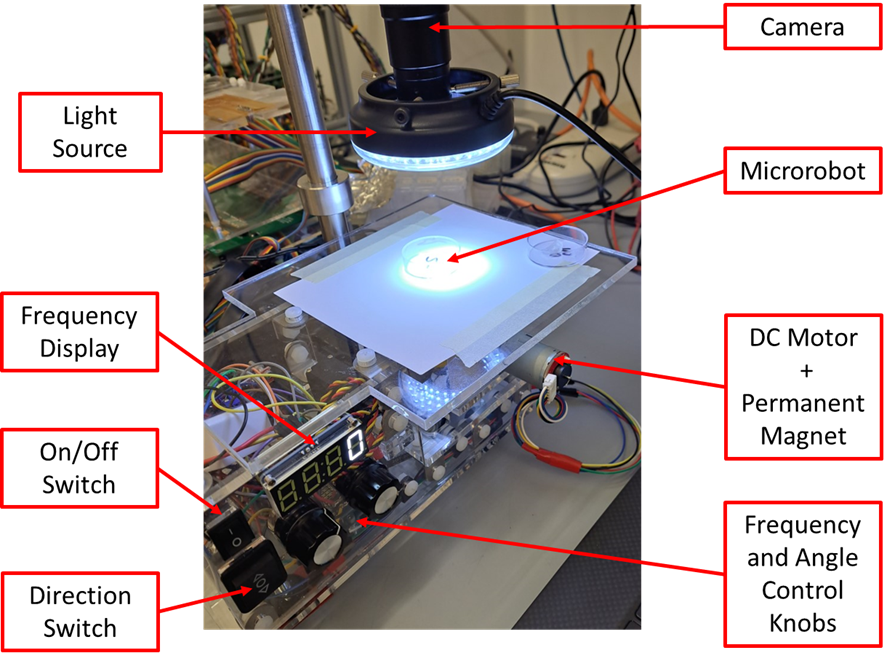}
\caption{Overview of the rotating magnetic field actuation system (RMFAS). It has an operating workspace approximately 75 mm in diameter. 
It can generate rotating magnetic fields in frequencies up to 5 Hz with 360$^{o}$ control of the in-plane direction of the field.}
\label{fig:RMAS}
\end{figure}

To propel the $\mu$TUMs, we employ a Rotating Magnetic Field Actuation System (RMFAS), as illustrated in Fig.~\ref{fig:RMAS}. This system utilizes a permanent magnet attached to two motors—one for out-of-plane rotation and the other for in-plane direction control of the microrobot. In contrast to electromagnetic systems like the Magnebotix MFG100 or Helmholtz coils systems often used~\cite{Shao2021}, our RMFAS uses a permanent magnet to provide a consistent high strength field \hl{(10-30mT)} enabling a large workspace without encumbering ultrasound-based imaging equipment. 
The magnetic field rotation rate is modified to observe the relationship to the velocity of the $\mu$TUM. The RMFAS provides a robust and consistent tool for reliably changing these parameters (rotation rate and direction of the magnetic field), keeping the results consistent throughout the tests with different microrobot types and test environments.

Real-time videos capture microrobot movements during each test. For \textit{in vitro} assessments, we employ a CMOS camera (Basler Ace2 - 45ucPRO) with a variable zoom microscope lens and ring light. For \textit{in vivo} locomotion testing, high-frequency ultrasound imaging (Vevo 3100, FUJIFILM VisualSonics, Toronto, ON, Canada) records microrobot videos. A custom Python script, utilizing an object tracking algorithm (CSRT), extracts data on the microrobot's position from the videos.

\subsubsection{\textit{In vitro} testing}

The \textit{in vitro} testing protocol established for evaluating the performance evaluation of the $\mu$TUMs is comprised of a comprehensive 9-panel examination. Each robot, of identical version, undergoes tripartite testing sessions, wherein its operational characteristics are assessed three times. The results are subsequently averaged to derive a final estimate for the average translational velocity ($V_{avg}$). This procedure is repeated in triplicate for every version of the \textmu TUMs to ascertain the average translational velocity ($V_{avg}$) on a flat surface across actuation frequencies of 2 Hz, 3 Hz, and 5 Hz.
Furthermore, the maximum incline angle ($\theta_{max}$) capabilities of each $\mu$TUM were evaluated using a progressive pass-fail system, incrementally increasing the incline by 5 degrees with each successful passage. The frequency was held constant at 5 Hz to ensure uniform results among varying incline adjustments. To emulate real-world biological conditions, the inclines were coated with Gelatin $\#$0 (Humimic Medical) sheets, enhancing surface friction and simulating more accurate climbing dynamics. The 9-panel testing methodology was similarly employed for assessing the maximum incline angle ($\theta_{max}$) performance. For each type of test, wet and dry environments were considered with repetition of the aforementioned testing regimen in each condition. Dry testing was performed on polymethacrylate (Formlabs Black FLGPBL04) 3D printed substrates in air. Wet testing was done under DI water on Gelatin \#0 substrates.

\subsubsection{In phantom testing}
The phantom of the distal colon was placed onto the magnetic actuation stage and filled with saline. The robot was then placed into the phantom. A small amount of ultrasound gel was placed on the phantom. Data was collected with actuation frequencies of 2 Hz, 3 Hz, 4 Hz, and 5 Hz utilizing high-frequency ultrasound microrobot tracking.

\subsubsection{\textit{In vivo} testing}
\label{sec:in_vivo_testing}
\begin{figure}[ht]
    \centering
    \includegraphics[height=2.5in]{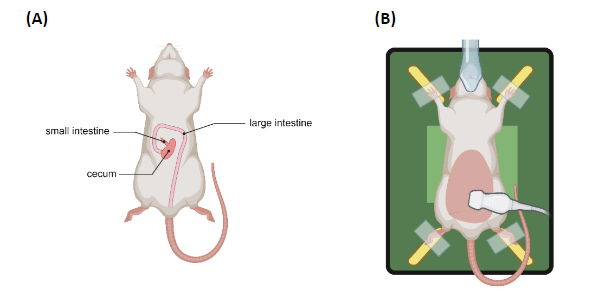}
    \caption{In vivo testing: (A) Schematic of the colon geometry superimposed on a rat. (B) Schematic of an anesthetized Sprague-Dawley rat on the Vevo 3100 ultrasound stage for rodents. Image created in BioRender.com.}
    \label{fig:invivoSchematic}
\end{figure}

Methods of animal use were compliant with the Institutional Animal Care and Use Committee at Purdue University (protocol 2002002016). Sprague-Dawley female rats (n=2; 1.5 years of age) were used for multiple imaging sessions. To prepare for ultrasound imaging (Vevo3100 Imaging system, FUJIFILM VisualSonics, Inc., Toronto, ON, Canada), the animals were fasted for approximately three hours. The rats were lightly anesthetized using 1-3\% isoflurane at a flow rate between 0.8 and 1.0 L/min. The animal was placed on the warming stage in the supine position while maintaining anesthesia with inhaled isoflurane (2\% in medical grade air) via a nose cone, as seen in Figure \ref{fig:invivoSchematic}. This warming stage was securely placed above the magnetic actuation stage. The animals were secured in place by taping the paws on electrodes that monitored the animal’s physiological signals, including heart rate, respiration rate, and temperature. To prepare the abdomen for imaging, the area was first shaved followed by application or topical depilatory cream to remove any remaining fur.

To prepare the animal for microrobot insertion, we first inserted a 3 mL syringe with a small hollow tube attached with a luer connection filled with sterile ultrasound gel into the rectum of the animal. We then inflated the colon with ultrasound gel working proximally to distally to remove the remaining stool in the distal portion of the colon. The lumen of the colon was then flushed with normal saline to remove the remaining ultrasound gel. The robot was then placed at the opening of the anus and inserted with the end of the tube, and the anus was gently closed by cross-clamping clothespins. Then the colon was inflated using saline to allow the robot to tumble within the lumen.

Once the microrobot was inserted and the anus was securely closed, a generous amount of ultrasound gel was applied along the animal’s midline and the ultrasound probe was lowered onto the animal to locate the starting position of the robot. Once located, the magnetic stage was activated to assess robot locomotion at the following speeds: 2 Hz, 3 Hz, 4 Hz, and 5 Hz. Locomotion was tracked in the proximal and distal colon, but data was assessed only in the distal colon to allow for consistency of data collection.\hl{Upon the conclusion of \textit{in vivo} testing, the clamp on the anus is released. This, coupled with natural peristaltic motion, causes the microrobot to be expelled from the colon in most cases. Repeating the saline flush performed prior to insertion of the microrobot can also induce expulsion of the microrobot.
}

\subsection{Biocompatibility testing}

Short-term cytotoxicity of the $\mu$TUM robot chassis, along with nickel-coated magnets were assessed. The $\mu$TUM robots were printed using the Formlabs clear general-purpose resin (Formlabs FLGPCL04). NIH3T3 murine fibroblasts were seeded in direct contact with the printed resin and nickel-coated magnets. We evaluated both whole and crushed magnets, and studied the metabolic activity over the course of three days. Cell proliferation was assessed using fluorescence microscopy (BioTek Cytation 5 Cell Imaging Multi-Mode Reader).
For cell viability studies, NIH3T3 murine fibroblasts were also seeded on negative and positive controls consisting of cells cultured in media and cells cultured in 70\% ethanol, respectively. At 24 and 48 hours, a PrestoBlue (Invitrogen PrestoBlue HS Cell Viability Reagent) assay was done to determine cell viability. After each assay, the diluted PrestoBlue solution was removed, and fresh media was replaced for cells to continue growing. Cell viability was also examined and assessed using fluorescence microscopy (BioTek Cytation 5 Cell Imaging Multi-Mode Reader).

\subsection{Drug Release Characterization Testing}

\subsubsection{Wax Formulation Melting Point Testing}\label{sec:WaxFormulationMethod}

The thermally sensitive wax coating for the $\mu$TUMs was created with a mixture of paraffin wax and mineral oil. By changing the mass fraction of mineral oil, the melting point of the wax mixture can be tuned for \textit{in vivo} drug release.   The optimal concentration should melt between 38\degree C and 42\degree C. At this range, there is no thermal damage to healthy intestinal tissues~\cite{Liang2015a,Draper2010a}, a low risk of denaturing the drug solution, and a low chance of unwanted release at body temperature, approximately 37\degree C. Eight different formulations, based on the mass fraction of mineral oil ($w$), were tested to identify the most desirable combination for drug release. \hl{The mass fractions that were tested ranged from $w$ = 0.1 to $w$ = 0.8 in increments of 0.1.}
Each wax formulation is placed in its solid form in a Petri dish which is then placed on top of a crystallization Pyrex dish with MilliQ water at 30\degree C. The waterbed is heated at approximately 0.01\degree C/s until the sample has completely melted using a hot stir plate (ThermoScientific Cimarec Digital Stirring Hotplates, 4.25 x 4.25 inches). The water temperature is monitored using a digital thermal probe. Two melting point values are recorded for each sample: the initial (when sample begins to melt) and final melting point (when sample has fully melted). The entire melting process per sample is recorded to further verify the reported melting point values.

\subsubsection{BSA Release Testing}
\begin{figure}
     \centering
     \includegraphics[width= 6 in]{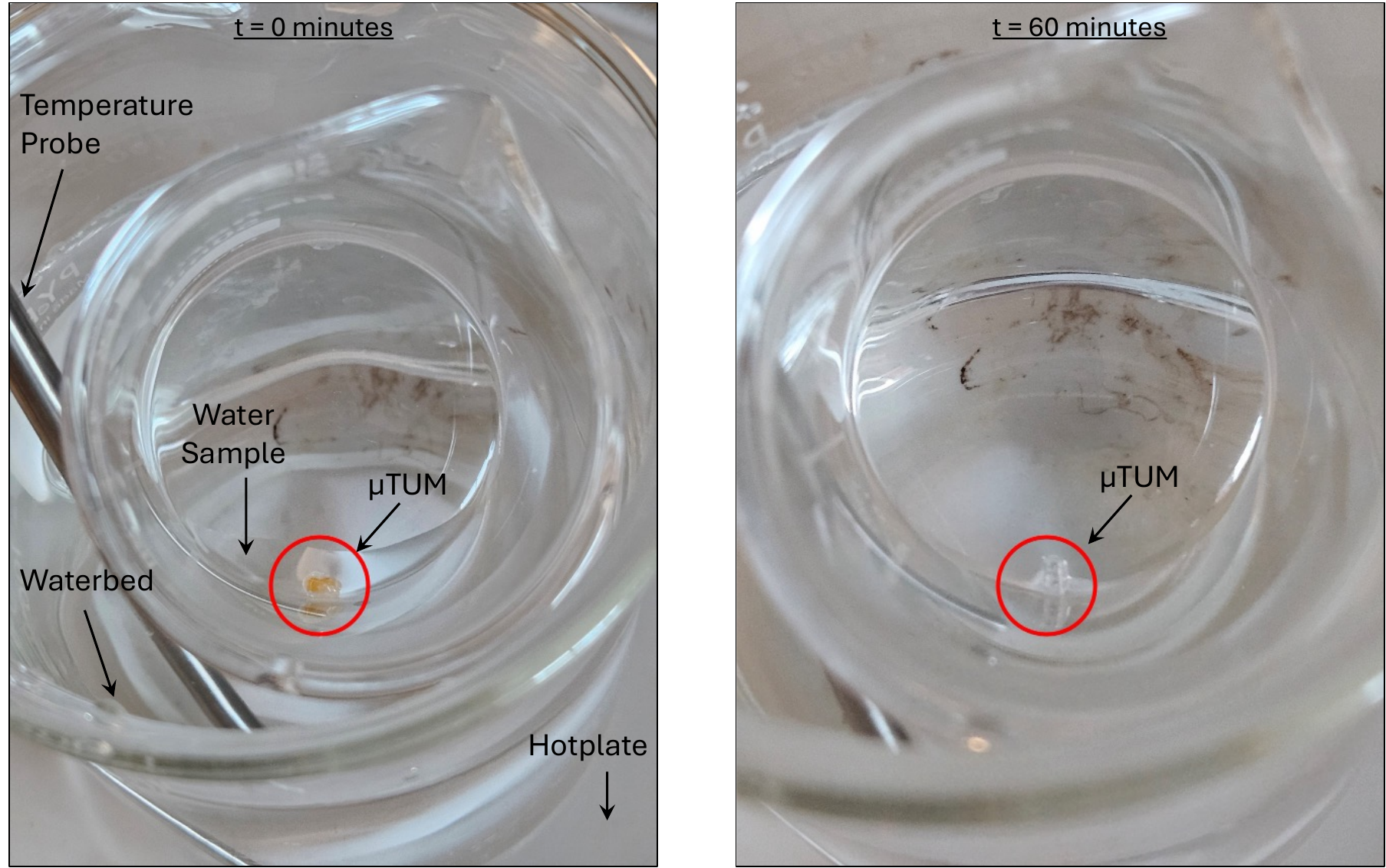}
     \caption{Experimental setup for BSA release study at t = 0 min and t = 60 min, respectively.}
     \label{fig:BSARelease}
 \end{figure}

\begin{figure}
    \centering
    \includegraphics[height=0.3\textheight]{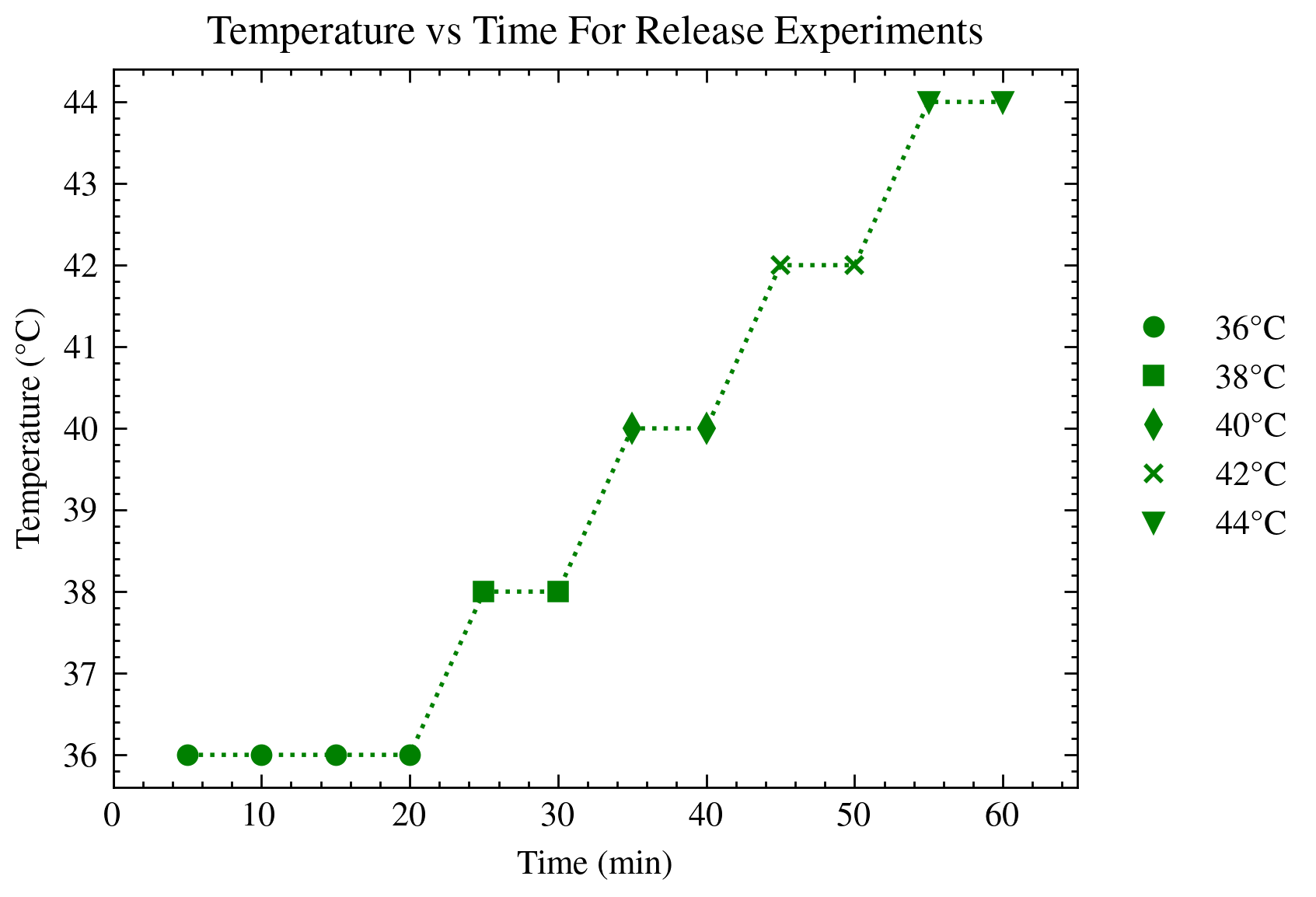}
    \caption{Graphical representation of how the temperature was changed during the BSA release study. Body temperature (circle: 36\degree C) is held for 20 minutes to verify no premature release takes place. Every temperature afterward is held for 5 minutes to test if release takes place during or after the temperature transition. 
    }
    \label{fig:TempPlot}
\end{figure}

$\mu$TUM-TP robots were loaded with a 100mg/mL fluorescent BSA (Albumin from Bovine Serum (BSA), Alexa Fluor 488 conjugate) solution and coated with a $w$= 0.6 wax formulation. After the wax is solidified, the robot is placed in a 100 mL Pyrex beaker with 200\textmu L of MilliQ water. The robot remains still during the study. This beaker is placed on a waterbed on a hot plate (Torrey Pines Scientific HS40 Programmable Digital Stirring Hot Plate with Ceramic Top, 8 x 8 inches). The experimental setup for these tests is shown in Fig.~\ref{fig:BSARelease}. The temperatures tested (and corresponding times) in this study are shown in Fig.~\ref{fig:TempPlot}. At each data point, a sample is collected by extracting all 200\textmu L of MilliQ and replacing it with the same volume of MilliQ water at the same temperature from the waterbed. First, the water bed is heated to 36\degree C and held for 20 minutes while collecting samples every 5 minutes. This is to ensure that the robot is not releasing BSA prematurely. Then, the waterbed is heated to 38\degree C and a sample is collected. The temperature is held for 5 minutes before a sample is collected. The temperature is increased to 40\degree C and the process is repeated. This study ends after the second collection at 44\degree C, resulting in a total of 12 samples collected per robot. To quantify release, the absorbance of each sample was measured using a BioTek Cytation 5 Cell Imaging Multi-Mode Reader.

\subsubsection{In Phantom Tests}

\begin{figure}[ht]
    \centering
    \includegraphics[width=7.5in, height=2.5in]{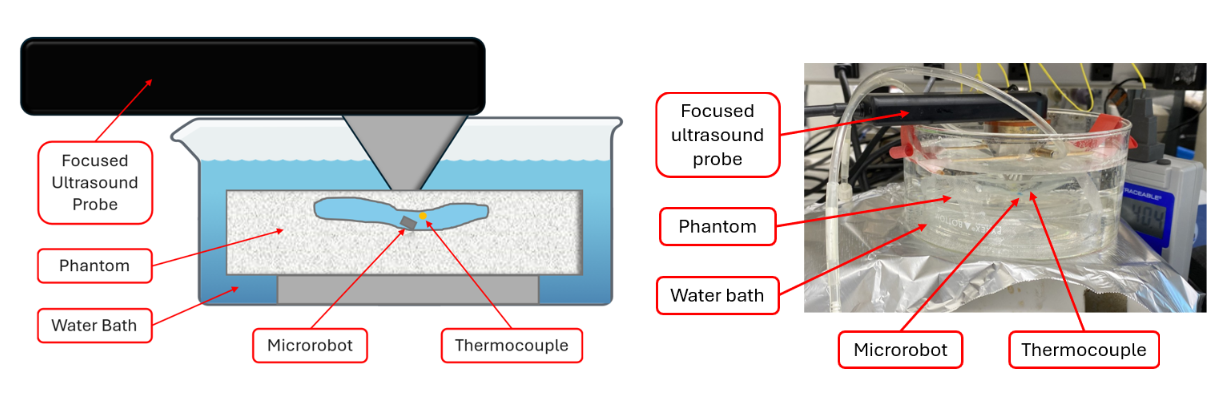}
    \caption{In phantom drug release testing. (Left) Schematic and (Right) actual image of phantom and focused ultrasound system setup. Left image created in BioRender.com.}
    \label{fig:phantomtestsetup2}
\end{figure}

For the \textit{in phantom} drug release tests, a large water bath is heated to 36°C using a hotplate. The phantom is then placed in this water bath to allow the warmed water to flow through the lumen of the phantom. The traceable waterproof type-k thermocouple (model: 15-078-186, 11789765) is used to monitor the temperature within the lumen throughout the procedure. The water bath is then moved to the magnetic stage and the robot is placed in the phantom using tweezers. The sonic concepts 33.0 mm x ROC 35 mm focused ultrasound transducer was placed in the water bath, above the phantom, and used to heat the water at the targeted location. The focused ultrasound settings were set to a power of 10 watts, frequency of 6.775 MHz, burst length of 0.20 ms, a period of 1.00 ms, and timed for 180 seconds. As the localized water temperature increases to 40-42$^{\circ}$C, the wax cap on the robot melts, allowing the dye to release. A schematic and image of the testing setup is shown in Fig.~\ref{fig:phantomtestsetup2}.

\section{Results and Discussion}

\subsection{Locomotion Characterization Testing}
\begin{figure}[ht]
    \centering
    \includegraphics[width=7.5in]{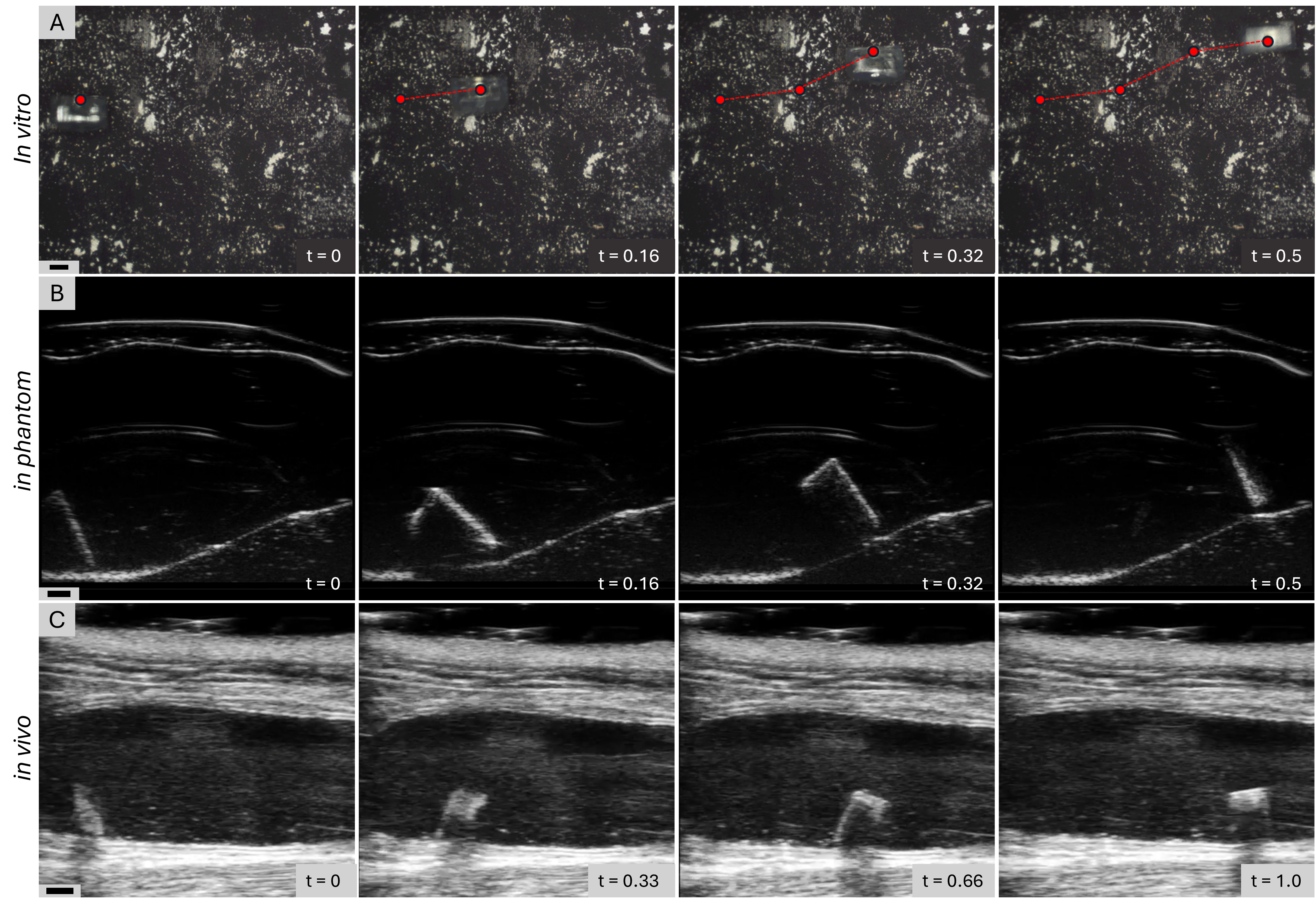}
    \caption{Locomotion testing.  Snapshots of $\mu$TUM locomotion in (A) \textit{in vitro}, (B) \textit{in phantom}, and (C) \textit{in vivo} environments. Note: Images from \textit{in vitro} tests are shown with an overhead view using a CCD camera.  The $\mu$TUM is traversing a gelatin\#0 sheet on a flat substrate.  Images for \textit{in phantom} and \textit{in vivo} tests are shown in a side view with imaging ultrasound. Scale bars = 1mm, t = time in seconds, frequency = 3Hz.
    }
    \label{fig:LocomotionTesting}
\end{figure}

Three different testing environments are used for the characterization testing.  They include \textit{in vitro}, \textit{in phantom}, and \textit{in vivo} rat testing environments.    
The primary objective of our \textit{in vitro} testing protocol is to measure the average translational velocity ($V_{avg}$) and maximum incline angle ($\theta_{max}$) for various $\mu$TUM geometries and testing variables. These variables encompass different $\mu$TUM designs and anticipated conditions during future use, such as diverse tumbling modalities, temperature induced drug release, and payload loaded/unloaded scenarios. Similar tests are then performed for the more realistic \textit{in phantom} test environment.  Finally, \textit{in vivo} tests are performed with as subset of conditions, based on the results from the \textit{in vitro} and \textit{in phantom} tests. The testing variables used for the characterization tests are shown in Table~\ref{tab:CharacterizationTestingParameters}. Snapshots from locomotion testing in the various test environments are shown in Fig.~\ref{fig:LocomotionTesting}.

\begin{table}
\centering
\caption{Locomotion characterization testing variables across test environments.}
\label{tab:CharacterizationTestingParameters}
\begin{tabular}{l|ccc|}
\cline{2-4}
 & \multicolumn{3}{c|}{\textbf{Test Environment}} \\ \hline
\multicolumn{1}{|c|}{\textbf{Test Variable}} & \multicolumn{1}{c|}{\textbf{In Vitro}} & \multicolumn{1}{c|}{\textbf{In Phantom}} & \textbf{In Vivo} \\ \hline
\multicolumn{1}{|l|}{\textbf{Locomotion Media}} & \multicolumn{1}{c|}{Dry, Wet: DI Water} & \multicolumn{1}{c|}{Wet: Saline} & Wet: Saline \\ \hline
\multicolumn{1}{|l|}{\textbf{Magnetic Frequency (Hz)}} & \multicolumn{1}{c|}{2, 3, 5} & \multicolumn{1}{c|}{2, 3, 5} & 2, 3, 5 \\ \hline
\multicolumn{1}{|l|}{\textbf{$\mu$TUM Geometry}} & \multicolumn{1}{c|}{TP, SP, EP} & \multicolumn{1}{c|}{TP, SP, EP} & TP \\ \hline
\multicolumn{1}{|l|}{\textbf{Temperature ($^{o}$C)}} & \multicolumn{1}{c|}{20, 50} & \multicolumn{1}{c|}{20} & 37 \\ \hline
\multicolumn{1}{|l|}{\textbf{Inclination Angle ($^{o}$)}} & \multicolumn{1}{c|}{0 to 50} & \multicolumn{1}{c|}{0} & N/A \\ \hline
\multicolumn{1}{|l|}{\textbf{Payload}} & \multicolumn{1}{c|}{Empty, Filled} & \multicolumn{1}{c|}{Empty, Filled} & Empty, Filled \\ \hline
\end{tabular}
\end{table}

\subsubsection{\textit{In vitro} Testing}
\begin{figure}
    \centering
    \begin{subfigure}{0.49\linewidth}
        \centering
        \includegraphics[width=\linewidth]{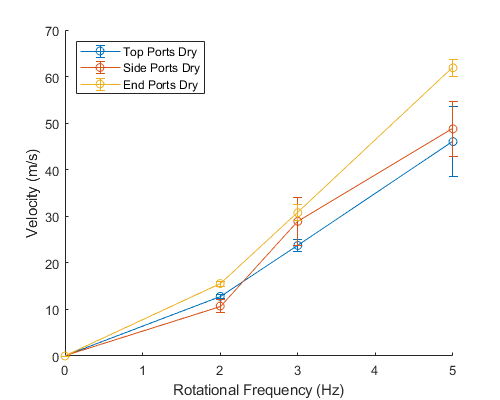}
        \caption{Dry environment velocity measurements}
        \label{fig:DryVelocity}
    \end{subfigure}
    \hfill
    \begin{subfigure}{0.49\linewidth}
        \centering
        \includegraphics[width=\linewidth]{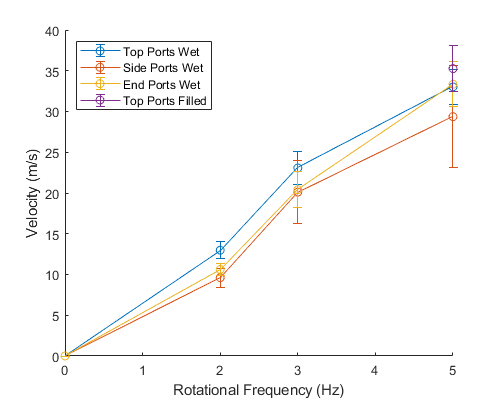}
        \caption{Wet environment velocity measurements}
        \label{fig:WetVelocity}
    \end{subfigure}
    \caption{\textit{In vitro} Locomotion Testing. Velocity of $\mu$TUMs vs magnetic actuation frequency in wet and dry environments with 0$^{\circ}$ slope. In the dry environment, $\mu$TUMs are placed on a 3D printed substrate of the same material as the $\mu$TUMs. In the wet environment, $\mu$TUMs are submerged in DI water on a Gelatin\#0 substrate.}
        \label{fig:Velocity}
\end{figure}

 The \textit{in vitro} velocity testing results are shown in Fig.~\ref{fig:Velocity}.  In both wet and dry environments, the $\mu$TUM velocity is roughly linearly related to the rotational frequency with only slight variance between robot types. The payload filled $\mu$TUMs exhibit comparable results to the empty $\mu$TUMs. Maximum climbable incline slope results are reported in Table~\ref{tab:slope}.  In each case, the different $\mu$TUM types were able to climb up to 20$^{\circ}$ slopes and 50$^{\circ}$ slopes in dry and wet environments, respectively.

\begin{table}
    \caption{Maximum climbable slopes for $\mu$TUM designs}   
    \centering
        \begin{tabular}[tbp]{ccc}
        \hline
                    &  Dry Environment  & Wet Environment\\\hline
        Top Ports   &  20\textdegree    & 50\textdegree\\
        Side Ports  &  20\textdegree    & 50\textdegree\\
        End Ports   &  20\textdegree    & 50\textdegree\\\hline
        \label{tab:slope}
    \end{tabular}
\end{table}

\subsubsection{\textit{In Phantom} Testing}\label{sec:InPhantomLocomotionResults}
\begin{figure}
    \centering
    \includegraphics[height=2.5in]{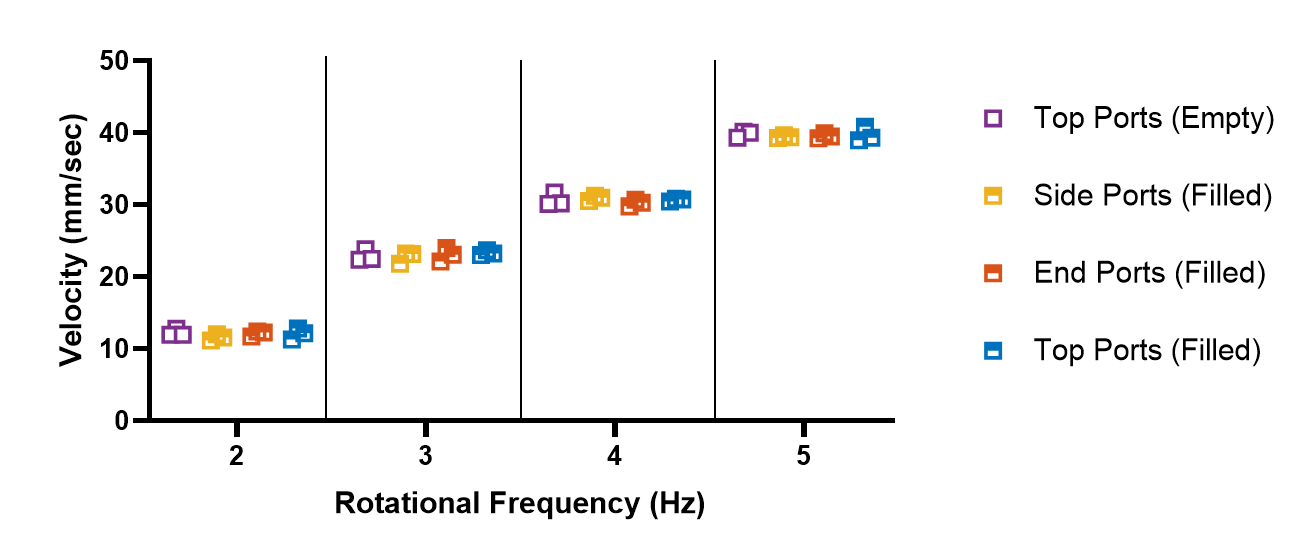}
    \caption{\textit{In phantom} Locomotion Testing. Velocity of various $\mu$TUM robots operating in the phantom vs magnetic actuation frequency.}
    \label{fig:PhantomVelocity}
\end{figure}

A similar set of locomotion tests were performed in a custom-made rat colon phantom made from Gelatin $\#$0 material filled with saline solution.  Here, we focused on testing each $\mu$TUM type across different magnetic actuation frequencies, in both the payload filled and empty states.  These results are plotted in Fig.~\ref{fig:PhantomVelocity}.  The velocities increase linearly with the actuation frequency, as expected.  It is noted here that the port location and geometry for the different $\mu$TUM types as well as if they are payload filled or empty, does not affect the $\mu$TUMs locomotion performance. The testing comprised of 3 replicates and 3 repeat measurements at each actuation frequency.


\subsubsection{\textit{In Vivo} Testing}

\begin{figure} 
    \centering
    \includegraphics[height=2.5in]{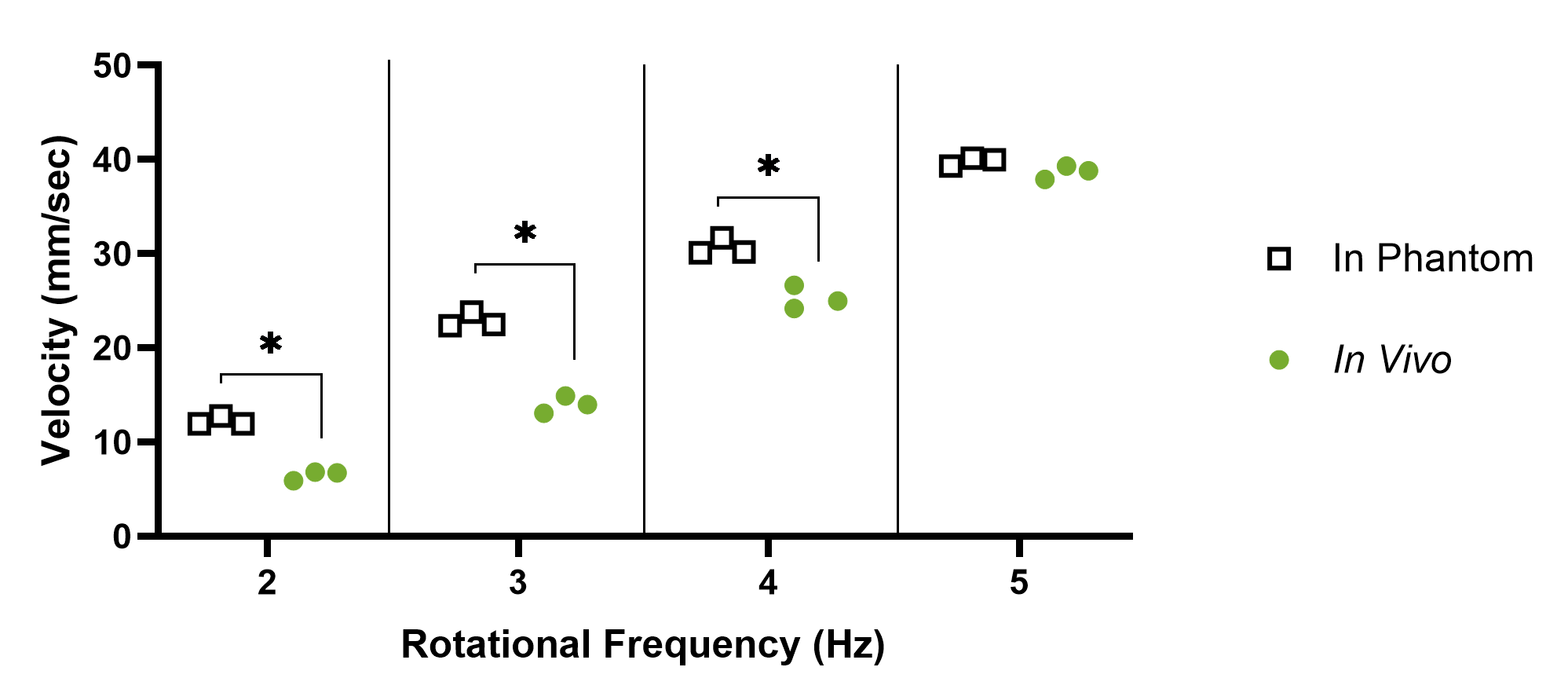}
    \caption{\textit{In vivo} Locomotion Testing. Velocity of empty $\mu$TUM-TPs \textit{in vivo} and in the phantom vs magnetic actuation frequency. *p \textless 0.05 indicating significant differences between \textit{in phantom} and \textit{in vivo} velocities at lower operating frequencies.}
    \label{fig:AnimalVelocity}
\end{figure}

\textit{In vivo} testing with Sprague-Dawley female rats was conducted.  Based on the results from the \textit{in phantom} testing (Sec.~\ref{sec:InPhantomLocomotionResults}) and the drug release study results (Sec.~\ref{sec:DrugLoadingResults}), the TP configuration ($\mu$TUM-TP) was selected in the payload empty configuration and tested across multiple magnetic actuation frequencies.  The results were then compared to the \textit{in phantom} results, as shown in Fig.~\ref{fig:AnimalVelocity}. The $\mu$TUM-TP robots behave similarly in both conditions at higher frequencies. However, we see that the \textit{in vivo} results are significantly slower than in the phantom at 2, 3, and 4 Hz. For each actuation frequency, 3 replicates and 3 repeat measurements were tested.
To evaluate significance of the results, the normality assumption was assessed under a Shapiro-Wilk test, which had a p-value (*p) less than 0.05. Homoscedasticity was assessed under Levene’s test. Since the data was not normal or log-normal, the parameters were analyzed using a Kruskal-Wallis with post-hoc Dunn test pairwise comparisons. Parameters were considered significant when the *p was less than 0.05.

\subsection{Biocompatibility Testing}
\begin{figure}[ht]
    \centering
    \begin{subfigure}[t]{0.49\linewidth}
        \centering
        \includegraphics[width=\linewidth]{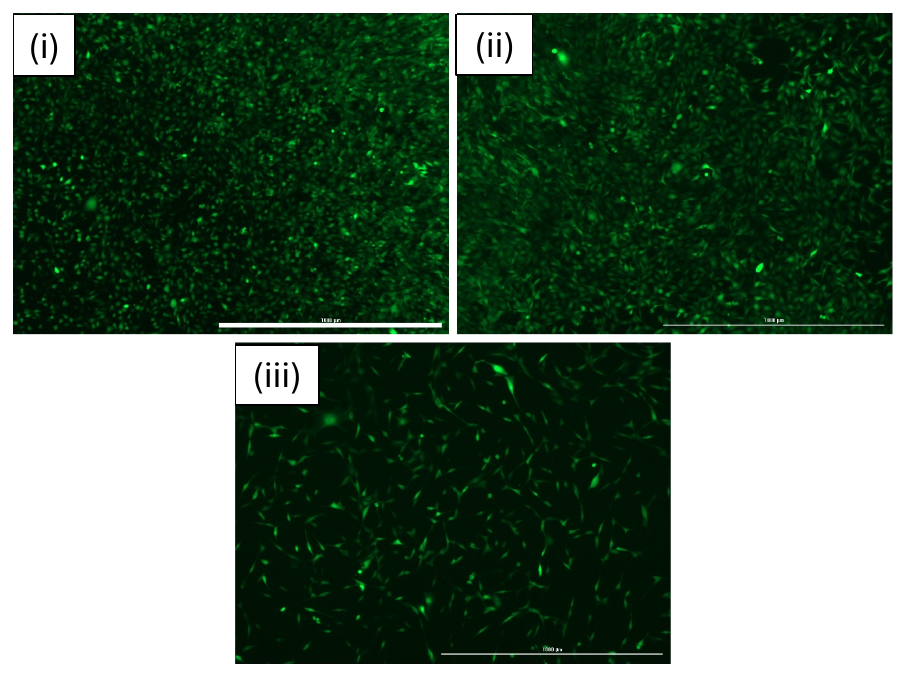}
        \caption{}
    \end{subfigure}
    \hfill
    \begin{subfigure}[t]{0.49\linewidth}
        \centering
        \includegraphics[width=\linewidth]{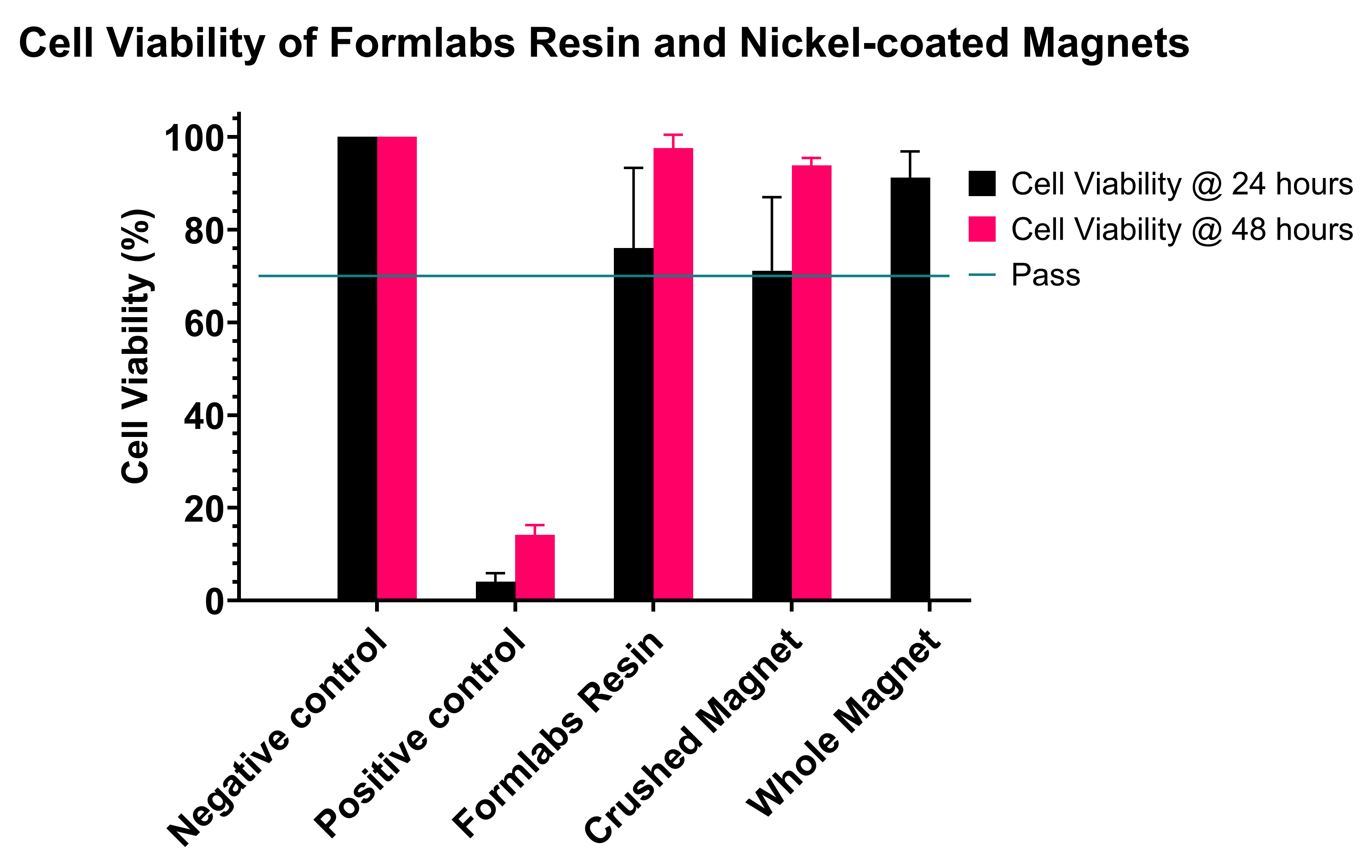}
        \caption{}
    \end{subfigure}
    \caption{Biocompatibility testing results. (a) Cell proliferation of different materials: (i) Formlabs resin; (ii) Crushed nickel-coated NdFeB magnet; (iii) Whole nickel-coated NdFeB magnet. Scale bar for all images are 1000 $\mu$m.  (b) Cell viability results for different materials along with negative and positive controls.}
    \label{fig:Biocompatibility}
\end{figure}

Short-term cytotoxicity of the assembled $\mu$TUMs and individual nickel-coated NdFeB magnets were assessed. Figure~\ref{fig:Biocompatibility}(a) illustrates the cell proliferation on the Formlabs general purpose clear resin used to 3D print the $\mu$TUM chassis and the nickel-coated magnets (whole and crushed) that are inserted into the $\mu$TUM chassis.  When seeded with NIH3T3 murine fibroblasts, all materials used in the composition of the $\mu$TUM demonstrated cell proliferation.
Figure~\ref{fig:Biocompatibility}(b) shows the cell viability of all experimental groups. According to ISO 10993-5, anything above 70\% is considered passing. While the viability of the whole nickel-coated magnets is well above 70\%, the viability of the Formlabs general-purpose clear resin and crushed nickel-coated magnets at 24 hours (black bars in Fig.~\ref{fig:Biocompatibility}(b)) are approximately 76\% and 71\%, respectively.  
Note: the crushed nickel-coated magnets are the worst case scenario should a magnet break or fracture once inside the biological space, which is very unlikely due the robot design and operating conditions.  For additional assessment, 
the cell viability at 48 hours (pink bars in Fig.~\ref{fig:Biocompatibility}(b)) was also examined for the Formlabs resin and crushed nickel-coated magnets. At 48 hours, the cytotoxicity for both the Formlabs resin and crushed nickel-coated magnets are above 70\%, at 94.77\% and 92.22\%, respectively.
When compared to the positive control, the cell viability of the Formlabs resin and the whole and crushed nickel-coated magnets were statistically significant at a p-value of less than 0.0001.  

\subsection{Drug Release Characterization Testing}\label{sec:DrugLoadingResults}

\subsubsection{Wax Formulation Melting Point Tuning}
To  create a thermally responsive release layer for the $\mu$TUMs, a wax mixture was made by combining paraffin wax and mineral oil. Pure paraffin wax has a melting point of approximately 50$^{\circ}$C. By changing the mass fraction of paraffin wax to mineral oil, the melting point can be reduced to a desired range between 38\degree C and 42\degree C that would allow for on demand delivery of the payload, at temperatures that would not cause thermal damage to the therapeutic agent or the local tissues.  
Eight different formulations, based on the mass fraction values of mineral oil ($w$), were tested to identify the combination to ensure on demand delivery of the therapeutic payload. 

\begin{figure}
    \centering
    \includegraphics[height=0.3\textheight]{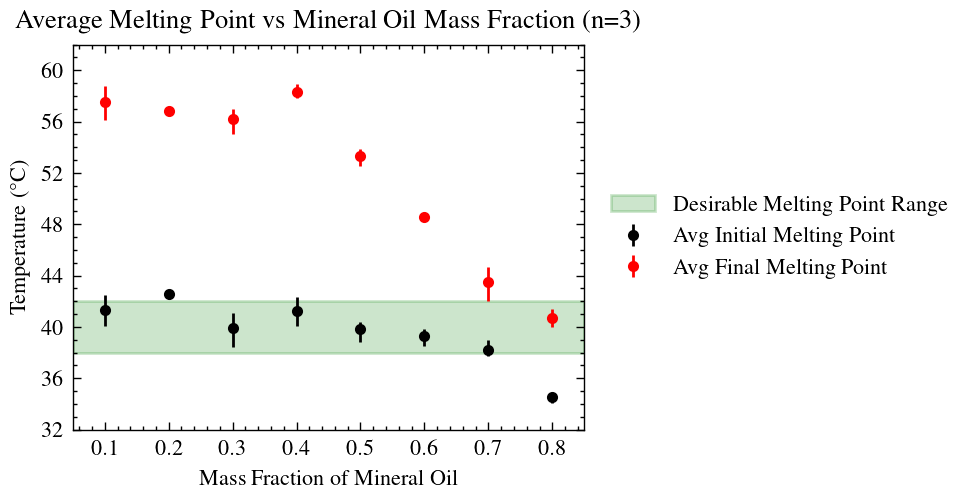}
    \caption{Wax Melting Point Study. Average initial and final melting point of each wax mixture (paraffin wax and mineral oil) with error bars representing maximum and minimum values measured at each mass fraction of mineral oil.}
    \label{fig:WaxMelting}
\end{figure}

The average initial and final melting point temperatures for each mass fraction values from $w$ = 0.1 to $w$ = 0.8\hl{, in increments of 0.1,} were recorded (n=3), as shown in Fig.~\ref{fig:WaxMelting}.
The average final melting temperature starts to decrease when  $w$ = 0.5 or greater.  
The solidified max mixture was noticeably softer for these $w$ values as well. Given the desirable melting point range which reduces the risk of damaging tissues, denaturing the protein solution, and prematurely releasing at body temperature, the optimal wax formulation is at $w$ = 0.7. However, at this high $w$ value, the wax mixture becomes very soft even at body temperature (37$^{\circ}$C), leading to the wax coating slipping from the robot and releasing its contents prematurely. Therefore, for the release study, a value of $w$ = 0.6 was selected. With $w$ = 0.6, the wax mixture will start to melt at approximately 39$^{\circ}$C.  This is acceptable since the drug release does not rely on the wax melting entirely, i.e., the drug will be released once the wax mixture starts to melt (the wax coating is significantly thin compared to the robot's dimensions). 

\subsubsection{\textit{In vitro} Drug Release Testing}
\begin{figure}
    \centering
    \includegraphics[width=2 in]{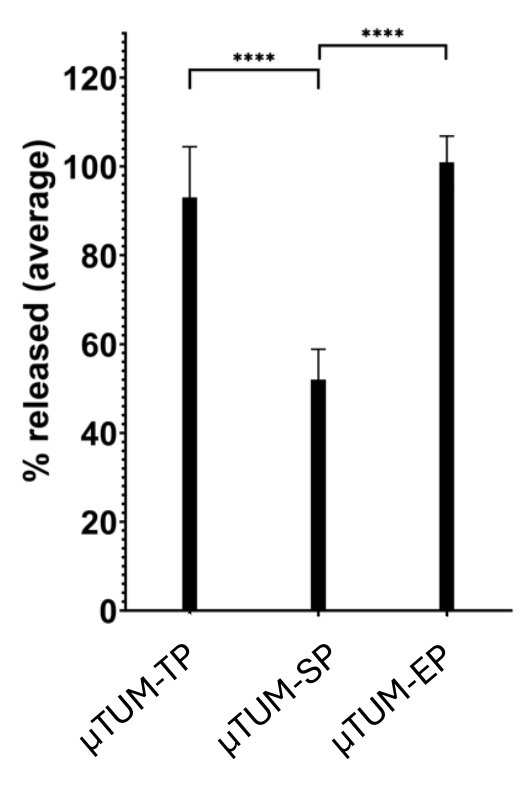}
    \caption{\hl{\textit{In Vitro} Drug Release Study. Average BSA release by percentage for different $\mu$TUM designs (n = 3) after local heating to 42$^{\circ}$C for 10 minutes.}}
    \label{fig:DrugReleaseRobotType}
\end{figure}

\hl{Initial \textit{in vitro} drug release tests were performed with each type ($\mu$TUM-TP, -SP, and -EP) of design, loaded with a 300mg/mL concentration of fluorescent BSA (Albumin from Bovine Serum (BSA), Alexa Fluor 488 conjugate) solution as a drug payload and capped with the tuned wax formulation. The loaded robots were placed into a beaker of PBS solution, heated to 37\degree C.  After 10 minutes, the temperature of PBS solution was heated to 42\degree C.  Release of the BSA was recorded after ten minutes at this temperature.  The results are shown in Fig.}~\ref{fig:DrugReleaseRobotType}.  \hl{The average percent BSA release across n = 3 samples for each robot type were 93\%, 52\%, and 100\% for the $\mu$TUM-TP, $\mu$TUM-SP, and $\mu$TUM-EP designs, respectively.  Thus, the percent BSA released was not statistically different between the $\mu$TUM-TP and $\mu$TUM-EP microrobots but is statistically significant compared to the $\mu$TUM-SP design, from on a Tukey HSD analysis of the data.  Based on these results, further drug release testing was performed using the $\mu$TUM-TP designs, as they are easier to load and cap than then $\mu$TUM-EP versions while exhibiting the same drug release characteristics.
}

\begin{figure}
    \centering
\captionsetup[subfigure]{justification=centering}
    \begin{subfigure}[t]{0.52\linewidth}
        \includegraphics[height=0.3\textheight]{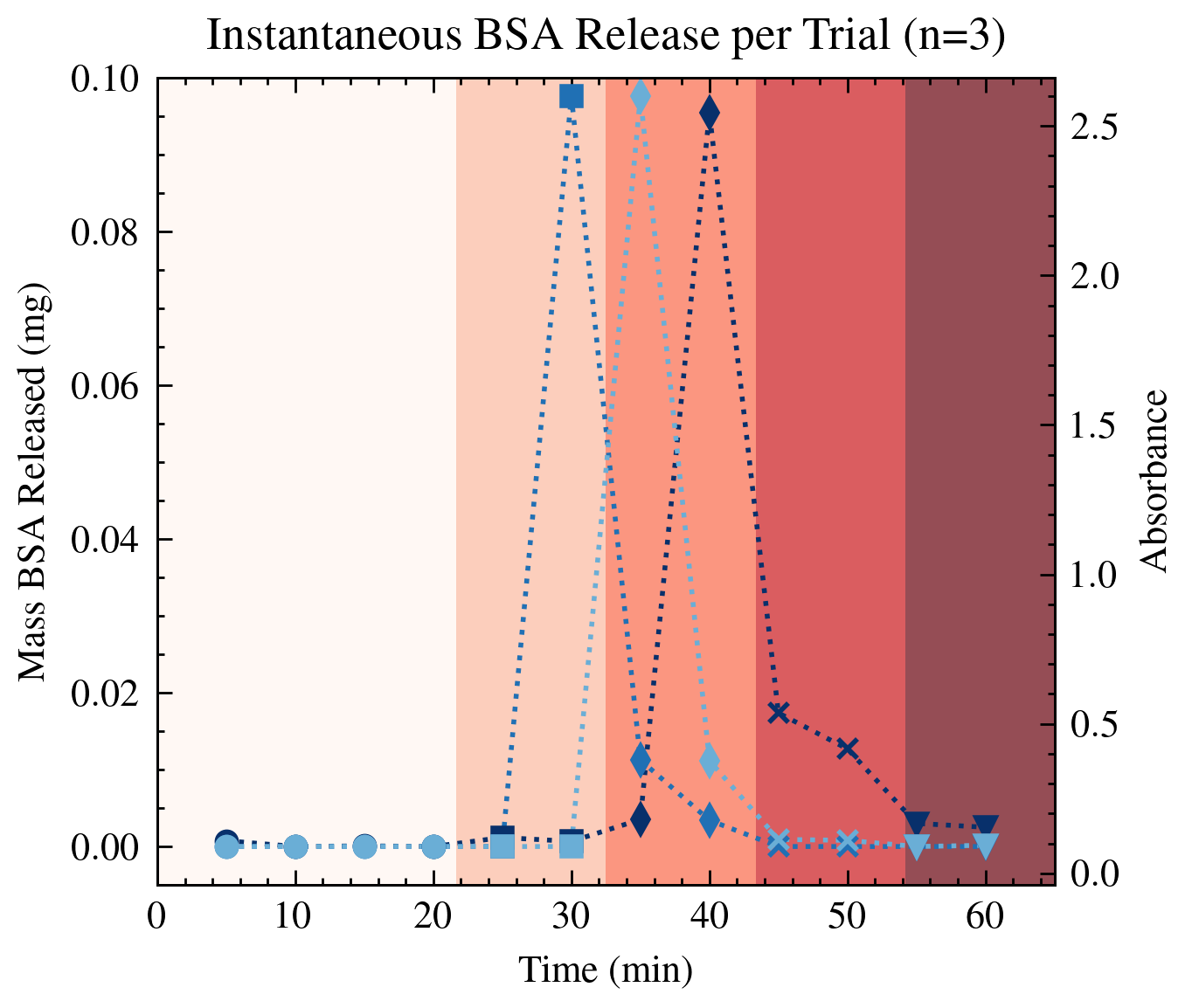}
        \caption{}
    \end{subfigure}
    \hfill
    \begin{subfigure}[t]{0.47\linewidth}
        \includegraphics[height=0.3\textheight]{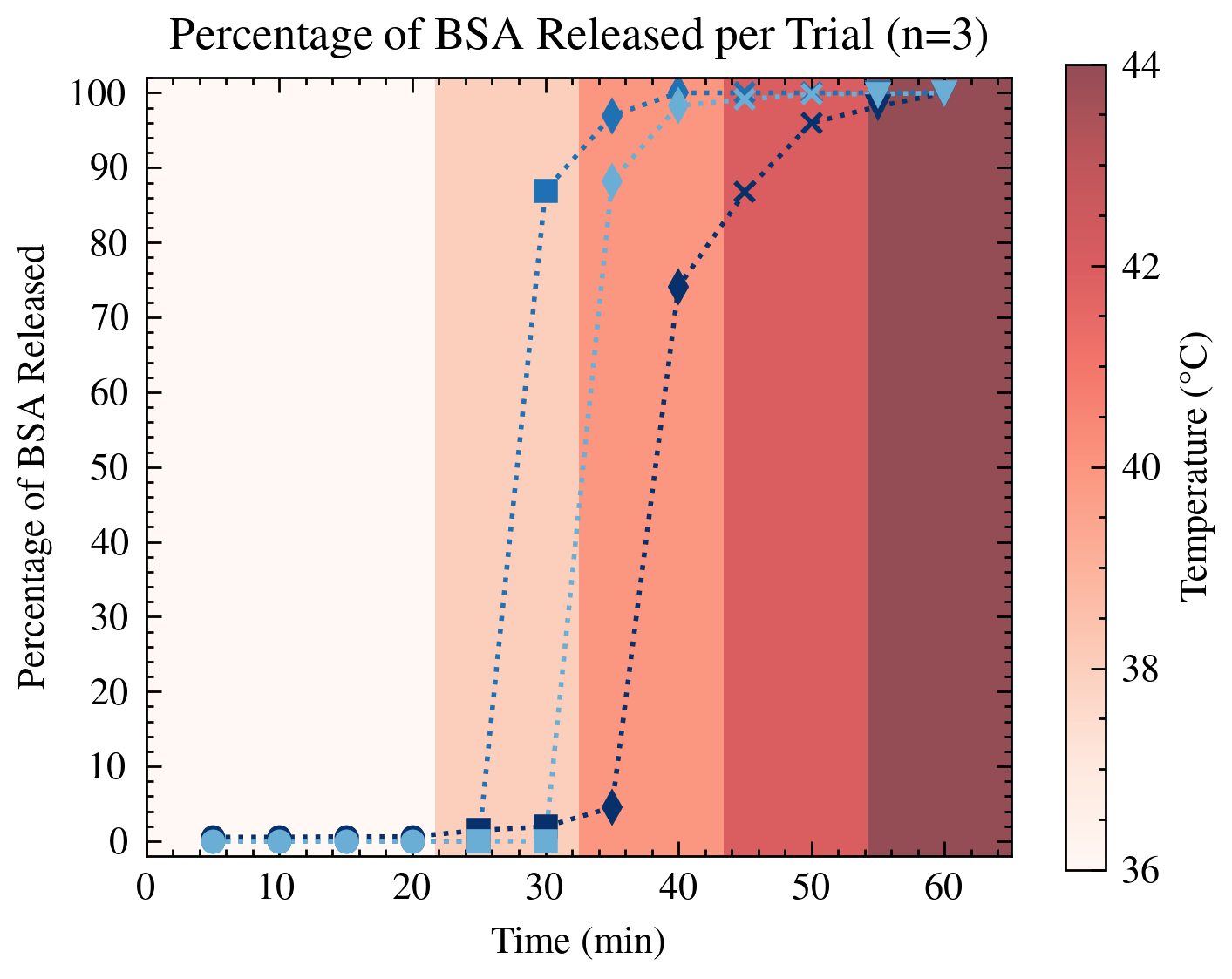}
        \caption{}
    \end{subfigure}
    \caption{\textit{In vitro} Drug Release Study. BSA Release from different $\mu$TUM-TP designs (n = 3): 
    (a) Mass of BSA released from the robot at each time point. (b) Percent of BSA released at each time point.}
    \label{fig:BSAPlots}
\end{figure}

\hl{For the next set of finer resolution \textit{in vitro} drug release tests,}
top port robots ($\mu$TUM-TPs) were loaded with a 100mg/mL fluorescent BSA (Albumin from Bovine Serum (BSA), Alexa Fluor 488 conjugate) solution as a drug payload and coated with a $w$ = 0.6 wax formulation. After the wax was solidified, the robot were placed in a temperature controlled water bath for testing.  The drug release results are shown in Fig.~\ref{fig:BSAPlots}.  Release of BSA was recorded at temperatures of 36$^{\circ}$C, 38$^{\circ}$C, 40$^{\circ}$C, 42$^{\circ}$C, and 44$^{\circ}$C.
There is no detectable release of the BSA protein at body temperature (37\degree C). A step-like release took place after holding 38\degree C for 5 minutes, during the transition between 38\degree C and 40\degree C, or after holding 40\degree C for 5 minutes (Fig.~\ref{fig:BSAPlots}(a)). All these trials exhibited release in the desirable range 
between 38\degree C and 42\degree C. Furthermore, all replicates released about 80\% of the BSA solution loaded into the robots (Fig.~\ref{fig:BSAPlots}(b)). These findings suggest that the robots can release drug payloads on demand with a controlled temperature increase. 

\hl{Note: considering current clinical methodologies for the treatment of IBD, a single dose of mesalamine (4g) is administered as a 60 mL enema (approximately 67 mg/mL) while a single dose prefilled pen of infliximab (120 mg) is administered as a 1 mL subcutaneous injection (120 mg/mL). Monoclonal antibody therapies, such as infliximab, that are administered subcutaneously, have doses that can be as high as 200 mg/mL.  These doses are required because of the substantial loss of antibody that occurs as it transports through the subcutaneous tissue en route to the lymphatics where it then becomes bioavailable. The viscosity of these formulations can range between 0.01 Pa*s to 0.33 Pa*s. Therefore, two different concentrations were used in this study, one at 300 mg/mL BSA because the viscosity approaches the 0.01 Pa*s commonly found in the subcutaneously administered formulations, and represents the worst case scenario for current formulations that could be directly translated for use within our delivery system. The 100 mg/mL concentration provides a lower viscosity formulation that still has a substantially high concentration that can be used to evaluate release and is readily available for use without modification.}

\subsubsection{\textit{In Phantom} Drug Release Testing}
\begin{figure}
    \centering
    \includegraphics[width=5in, height=2.5in]{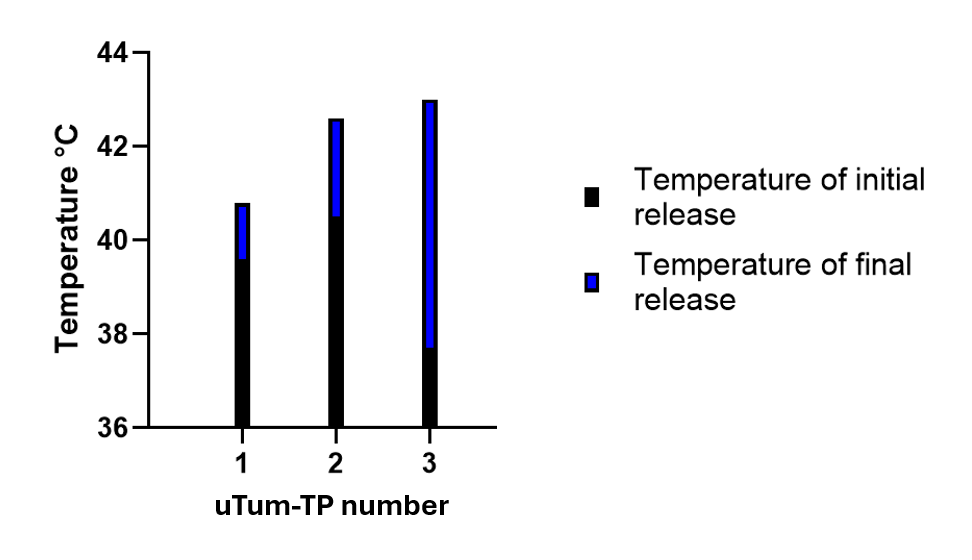}
    \caption{\textit{In phantom} Drug Release Study. Initial and final release temperatures for different $\mu$TUM-TP samples in the phantom (n = 3).}
    \label{fig:dyereleasegraph}
\end{figure}

\begin{figure}
    \centering
    \includegraphics[width=7.5in, height=2in]{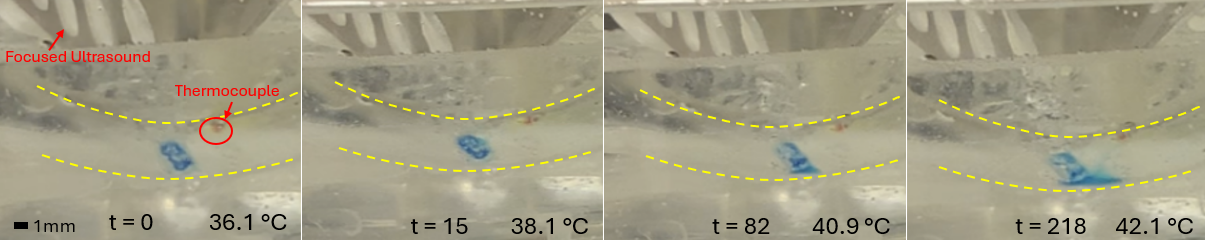}
    \caption{\textit{In phantom} Drug Release. Snapshot of dye release of $\mu$TUM-TP robot in the phantom using focused ultrasound for targeted heating. t = time elapsed in seconds. The upper and lower boundaries of the phantom can be seen outlined in the dashed yellow lines.}
    \label{fig:dyereleasesnaps}
\end{figure}

For the \textit{in phantom} drug release tests, a mock drug payload solution of blue food dye, 0.07\% Tween20, and MilliQ water was loaded into $\mu$TUM-TP robots.  
Once placed in the phantom, a focused ultrasound system was used to heat the robots and release the mock drug payload.  Results from these tests are shown in Figures ~\ref{fig:dyereleasegraph} and ~\ref{fig:dyereleasesnaps}, respectively.
A total of 3 robots were tested.
All of the trials had an initial release temperature above the average internal body temperature of 37$^{\circ}$C~\cite{ratcolontemp}.
In one case, the robot began its initial release of the payload at 37.7$^{\circ}$C. That robot continued to release the dye as the focused ultrasound warmed the region to around 42$^{\circ}$C. The total time to reach this final temperature was approximately 3-4 minutes. In another case, the robot began its initial release after 1.5 minutes at 40.9$^{\circ}$C (Fig.~\ref{fig:dyereleasesnaps}). Some residual dye remaining in the robot can be visually seen upon robot retrieval from the phantom. 

\section{Conclusion}
This work presents innovative designs for 3D-printed tumbling magnetic microrobots ($\mu$TUMs) specifically engineered for targeted \textit{in vivo} drug delivery. Three different $\mu$TUM design configurations were studied.  Each were realized using stereolithography to create the microrobot chassis.  The manual assembly of a micro-sized magnet into the chassis completes the assembly and allows for control with an external rotating magnetic field and tumbling locomotion. The materials comprising the $\mu$TUM designs proved to be biocompatible based on cell proliferation and cell viability studies. 
The paper outlines the design details for the different $\mu$TUM configurations and a comprehensive locomotion study is presented.  The $\mu$TUMs locomotion capabilities are evaluated in \textit{in vitro}, \textit{in phantom}, and \textit{in vivo} rat colon environments.  
A drug payload loading process was developed and a thermally sensitive wax coating formulation was tuned to achieve a desired melting point range for biomedical applications.  The wax coating formulation was tested on the $\mu$TUM top port configuration in \textit{in vitro} and \textit{in phantom} environments.  Local heating, via hotplate and a focused ultrasound system, respectively, demonstrated the controlled release of the drug payload from the $\mu$TUM.  These results showcase the potential for efficient and targeted drug delivery within the colon and large intestine using the $\mu$TUMs.  Future work will explore \textit{in vivo} targeted drug delivery in rat colons with various drug candidates and a study of therapeutic efficacy of the released drug.  

\medskip

\medskip

\medskip

%

\bibliographystyle{MSP}
\bibliography{library}

\end{document}